\documentclass[10pt,twocolumn,letterpaper]{article}
\usepackage[algorithms]{wacv}  
\usepackage[accsupp]{axessibility}
\usepackage{graphicx}
\usepackage{amsmath}
\usepackage{amssymb}
\usepackage{booktabs}

\usepackage{multirow}
\usepackage{graphicx}
\usepackage{xr}
\usepackage[dvipsnames,table,xcdraw]{xcolor}
\usepackage{float}

\usepackage{animate}
\definecolor{lightred}{rgb}{1,0.5,0.5}

\usepackage[pagebackref,breaklinks,colorlinks,bookmarks=false]{hyperref}

\usepackage[capitalize]{cleveref}
\crefname{section}{Sec.}{Secs.}
\Crefname{section}{Section}{Sections}
\Crefname{table}{Table}{Tables}
\crefname{table}{Tab.}{Tabs.}

\def\wacvPaperID{1690} 
\def\confName{WACV}
\def\confYear{2024}

\begin{document} 

\title{MuSHRoom: Multi-Sensor Hybrid Room Dataset for \\ Joint 3D Reconstruction and Novel View Synthesis}

\author{Xuqian Ren,$^{1}$ Wenjia Wang,$^{2}$  Dingding Cai,$^{1}$  Tuuli Tuominen,$^{1}$
 Juho Kannala,$^{3}$  Esa Rahtu$^{1}$  \\
{}$^1$Tampere University, Finland {}$^2$The University of Hong Kong, China {}$^3$Aalto University, Finland\\
{\tt\small \{xuqian.ren,dingding.cai,tuuli.tuominen,esa.rahtu\}@tuni.fi wwj2022@connect.hku.hk } \\
{\tt\small Juho.Kannala@aalto.fi }
}

\twocolumn[{
\renewcommand\twocolumn[1][]{#1}
\maketitle
\begin{center}
    \captionsetup{type=figure}
    \includegraphics[width=0.8\textwidth]{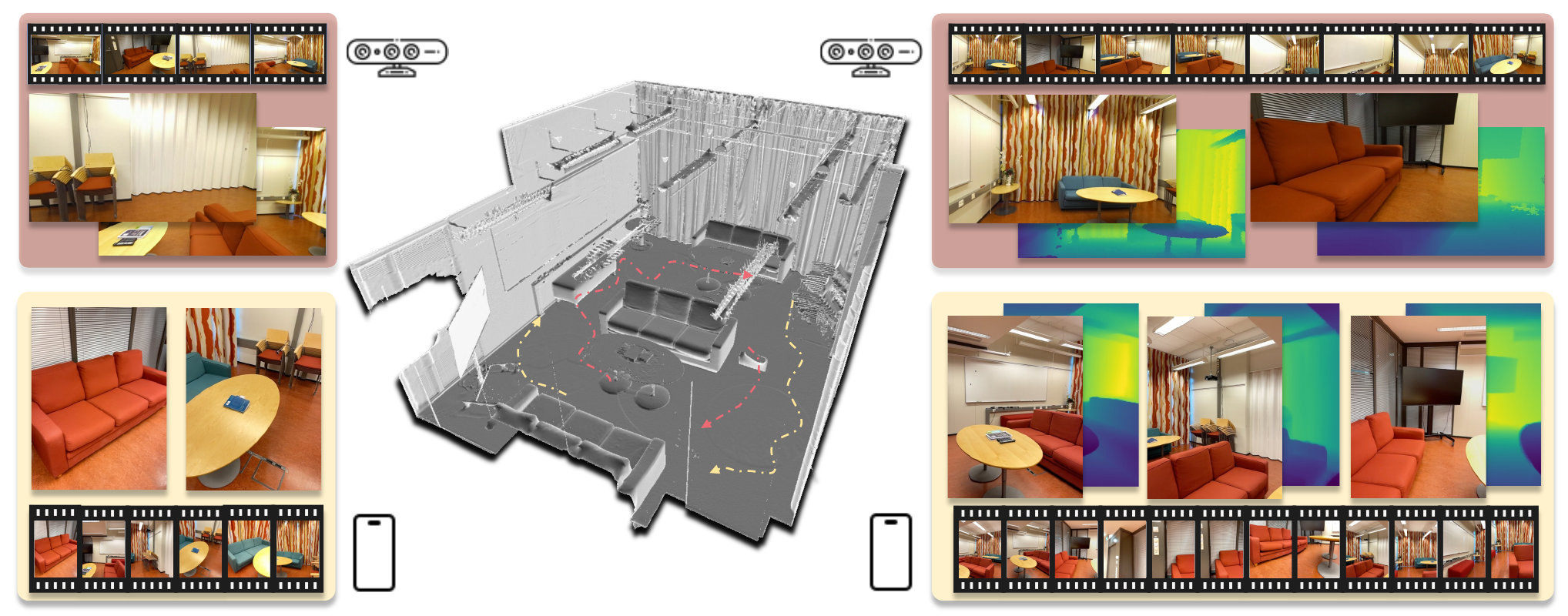}
    \captionof{figure}{The proposed MuSHRoom dataset includes 10 rooms captured by consumer devices Kinect and iPhone, and each room provides ground-truth mesh models obtained by a Faro scanner. Both Kinect and iPhone capture one long and one short RGB-D sequence for simulating a typical VR/AR use case. The MuSHRoom dataset provides camera poses and point clouds for Kinect and iPhone sequences. The dash lines demonstrate the rough capture trajectories. This dataset is intended for benchmarking room-scale 3D reconstruction and novel view synthesis.}
\end{center}
}]

\begin{abstract}
%
Metaverse technologies demand accurate, real-time, and immersive modeling on consumer-grade hardware for both non-human perception (e.g., drone/robot/autonomous car navigation) and immersive technologies like AR/VR, requiring both structural accuracy and photorealism.
However, there exists a knowledge gap in how to apply geometric reconstruction and photorealism modeling (novel view synthesis) in a unified framework. 
%
To address this gap and promote the development of robust and immersive modeling and rendering with consumer-grade devices, we propose a real-world \textbf{Mu}lti-\textbf{S}ensor \textbf{H}ybrid \textbf{Room} Dataset (MuSHRoom).
%
Our dataset presents exciting challenges and requires state-of-the-art methods to be cost-effective, robust to noisy data and devices, and can jointly learn 3D reconstruction and novel view synthesis instead of treating them as separate tasks, making them ideal for real-world applications. 
We benchmark several famous pipelines on our dataset for joint 3D mesh reconstruction and novel view synthesis.
Our dataset and benchmark show great potential in promoting the improvements for fusing 3D reconstruction and high-quality rendering in a robust and computationally efficient end-to-end fashion.
The dataset and code are available at the project website: \url{https://xuqianren.github.io/publications/MuSHRoom/}.

\end{abstract}    
\section{Introduction}
An effective way for artificial intelligence to understand and interact with the tangible realm is to simulate and extrapolate physical objects into a digital environment with the help of sensory input signals, such as RGB images or RGB-D images captured by cameras.
To realize the task of creating virtual representations of tangible entities, geometric reconstruction (3D reconstruction) and photorealism modeling (novel view synthesis, NVS) tasks have been proposed, and both of them play a significant role in the development of VR/AR~\cite{deng2022fov,huang2022stylizednerf}.
%
%
Nonetheless, current room-scale datasets do not support evaluating the two tasks jointly in a quantitative way, which hinders the state-of-the-art methods applied to VR/AR applications that require both geometry accuracy and photorealism.
%

%
%
%
Most of the current room-scale datasets~\cite{barron2022mip,tancik2023nerfstudio} either only contain RGB/RGB-D inputs without ground truth meshes for 3D reconstruction comparison, or are over-cleaned~\cite{Azinovic_Martin-Brualla_Goldman_Niebner_Thies_2022,wang2022go} and cannot fully reflect the challenges in the real world.
%
Redwood Scan Dataset~\cite{park2017colored} provides RGB-D inputs of real-world room scenes and industrial laser scans for mesh reference.
%
However, it only uses a single-capturing device in each room to capture a single sequence, which is not enough for simulating the real VR/AR use case. 
%

%

Considering the lack of proper benchmark and datasets, we propose a real-world \textbf{Mu}lti-\textbf{S}ensor \textbf{H}ybrid \textbf{Room} Dataset (\textbf{MuSHRoom}).
Our dataset focuses on indoor room-scale scenarios and raises interesting real-world challenges on occlusion, motion blur, reflection, transparency, sparseness, illumination diversity, etc.
Each room is captured with the Azure Kinect and iPhone consumer device for RGB-D sequences as inputs and an industrial laser scanner as geometry ground truth reference.
%
%
For each consumer device, we capture two sequences: one long capture with most of the details inside the room and another shorter sequence captured with an independent trajectory.
%

Based on the MuSHRoom dataset, we propose a new benchmark, aiming to evaluate both the reconstruction and NVS ability of methods.
Furthermore, we also propose a new protocol for practical NVS evaluation.
When evaluating NVS, previous methods~\cite{barron2022mip,tancik2023nerfstudio} usually uniformly sample frames from the whole sequences as the test set, which does not reflect the real case in VR/AR.
%
In our comparison protocol, we use the long sequences as the training set and the short sequences as the test set, which raises challenges in robustness since the camera positions and view directions have a large gap between these two captures when observing the same objects.
This evaluation protocol is common in AR/VR applications where the users will scan the whole room for the first time, and then VR glasses will render the reality according to the positions and view directions of users.

Most existing pipelines are designed to either perform excellent geometry modeling or photorealistic rendering. 
Based on the proposed MuSHRoom dataset and benchmark, we provide an extensive comparison of previous pipelines for both reconstruction and rendering quality.
The comparison also shows that the need for achieving both reconstruction and NVS tasks at the same time is clear and a long way.

Our contributions can be summarized as follows:
\begin{itemize}
    \item We make one of the first attempts to construct a dataset collected with multiple sensors for joint 3D reconstruction and novel view synthesis. We provide a detailed pipeline and program codes for capturing and processing the data, including information on the hardware setup, data acquisition, and post-processing steps. Our pipeline serves as a comprehensive guide for researchers interested in creating similar datasets.
    \item We provide an extensive comparison of off-the-shelf methods based on our new benchmark. Our evaluation provides insights into the strengths and limitations of each pipeline and their applicability to real-world scenarios.
    \item
     Our dataset raises new real-world challenges and practical evaluation protocol for the state-of-the-art methods to apply to real applications and encourages further exploration of the challenges and opportunities presented by our dataset.
\end{itemize}
\section{Related Work}

%
There are numerous datasets for the 3D reconstruction or novel view synthesis tasks. Therefore, we limit our discussion to the most related scene-level datasets and introduce benchmarks used for modeling and rendering.

\noindent\textbf{3D Room-Level Datasets.}
Redwood Scan~\cite{park2017colored} captures five real-world rooms with one single RGB-D camera and Faro scanner. It is the most similar dataset to ours, but it only uses one device to get the color and depth images.
%
%
%
Neural RGB-D Synthesis Datasets~\cite{Azinovic_Martin-Brualla_Goldman_Niebner_Thies_2022,wang2022go} are unified synthesized datasets that can be used for joint 3D reconstruction and novel view synthesis comparison.
%
%
To simulate real-world captures, noise and artifacts are manually added to the depth images, and BundleFusion~\cite{Dai_Nießner_Zollhöfer_Izadi_Theobalt_2016} is used to generate the estimated pose annotations.
However, real-world noise caused by motion blur, shaking, reflection, etc., cannot be easily simulated. Thus, the domain gap between these datasets and real scenes still remains.
%
%
%
%
ETH3D~\cite{schops2017multi} releases ground truth laser scans with registered images captured by multiple devices. However, they only provide high-resolution RGB images without depth, and the other device only provides grayscale low-resolution images.
ScanNet~\cite{Dai_Chang_Savva_Halber_Funkhouser_Niessner_2017} is targeted for 3D scene understanding. 
It contains a large volume of RGB-D sequences and is a valuable dataset for room-scale 3D reconstruction.
However, the evaluation can only be conducted qualitatively due to the lack of ground truth mesh models.
%
%

%

With the rapid prosperity of research on NeRF~\cite{mildenhall2021nerf}, new datasets are proposed for novel view synthesis.
Nerfstudio Dataset~\cite{tancik2023nerfstudio} contains object-scale and room-scale scenes captured with a mobile phone or mirrorless camera. 
Mip-NeRF 360 Dataset~\cite{barron2022mip} includes five outdoor scenes and four indoor scenes, among which only one sequence is captured in the room-scale scenario. 
Note that only RGB images are provided in the last two datasets.
Some datasets~\cite{Chang_Dai_Funkhouser_Halber_Niebner_Savva_Song_Zeng_Zhang_2017,Straub,Meuleman_Liu_Gao_Huang_Kim_Kim_Kopf_2023,Xiao_Owens_Torralba_2013,Baruch_Chen} are commonly used for scene reconstruction and rendering. However, they are either synthetic datasets~\cite{Chang_Dai_Funkhouser_Halber_Niebner_Savva_Song_Zeng_Zhang_2017,Straub} or lack ground truth meshes~\cite{Meuleman_Liu_Gao_Huang_Kim_Kim_Kopf_2023,Xiao_Owens_Torralba_2013}. 
\cite{Baruch_Chen} provides real RGB-D images with ground truth meshes but is only captured by a single device.


\noindent\textbf{3D Reconstruction and NVS Methods.}
Commercial software applications, such as Pixel4D~\cite{pixel4d} and Reality Capture~\cite{capturereality}, can be used for image-based reconstruction.
%
%
However, they require dense input sequences to guarantee precision and inevitably suffer significant performance degradation when the inputs are very sparse, limiting their applicability in room reconstruction with commercial devices.
Traditional methods like volumetric fusion~\cite{curless1996volumetric}, BundleFusion~\cite{Dai_Nießner_Zollhöfer_Izadi_Theobalt_2016}, KinectFusion~\cite{6162880} reconstruct 3D models from image sets based on geometric vision and graphics principles, but they are lack robustness of some complex scenes.
Based on volumetric rendering, Nerf++~\cite{Zhang_2020}, MipNeRF 360~\cite{barron2022mip}, Nerfstudio~\cite{tancik2023nerfstudio}, and zip-NeRF~\cite{barron2023zip} extend original NeRF~\cite{mildenhall2021nerf} to real-scene applications.
NICE-SLAM~\cite{Zhu_Peng_Larsson_Xu_Bao_Cui_Oswald_Pollefeys_2022} and NICER-SLAM~\cite{Zhu_Peng_Larsson_Cui_Oswald_Geiger_Pollefeys_2023} combine NeRF with simultaneous localization and mapping (SLAM) method, enabling real-time dense RGB-D SLAM system that can be applied to large-scale scenes.
NeuS~\cite{wang2021neus}, VolSDF~\cite{Yariv_Gu_Kasten_Lipman_2021}, Neural RGB-D~\cite{Azinovic_2022}, GO-Surf~\cite{wang2022go}, and BakedSDF~\cite{Yariv_2023} combine truncated signed distance function (TSDF) and volumetric rendering to minimize the geometry ambiguity.
Many of these pipelines prioritize either geometric accuracy or synthesis enhancement. This specialization can limit their effectiveness in VR/AR applications, which demand both structural accuracy and realistic immersion. 
%
\section{The MuSHRoom Dataset}
\label{sec:data_collection}
\begin{figure*}[t]
	\centering	 
	\includegraphics[width=1\linewidth]{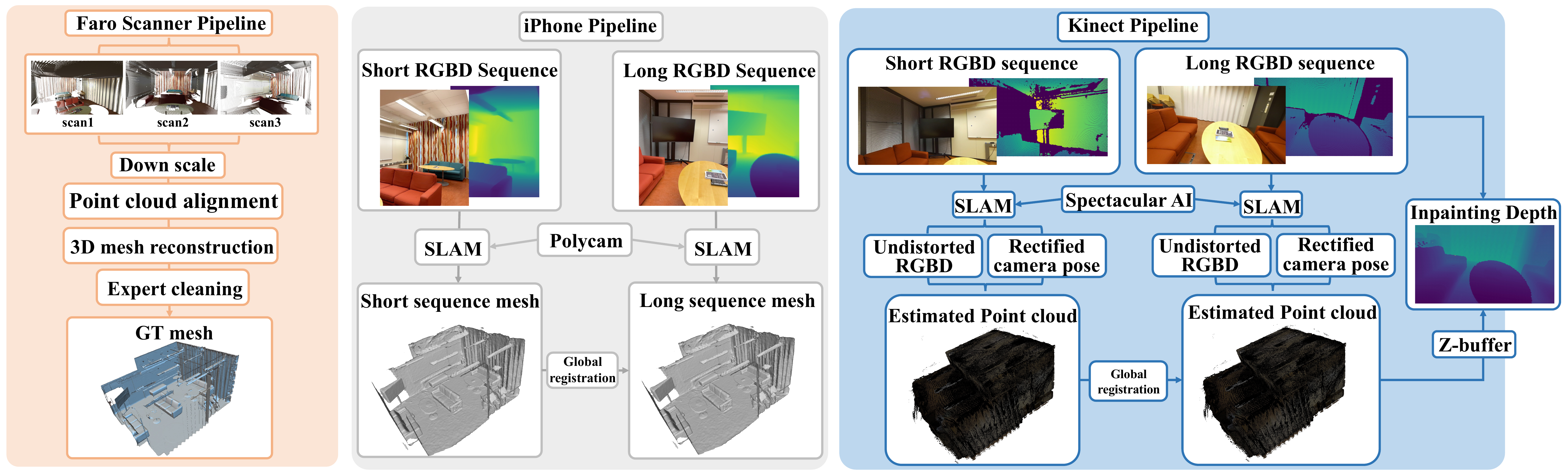}	 	
	\caption{The process pipeline. We use a Faro Scanner to obtain point clouds of the room from different locations and stitch them to create a complete model of the room, compensating for occluded areas. We use spectacular AI SDK to extract the undistorted RGB-D and camera pose for Kinect sequences and use the z-buffer to project point clouds into pixel coordinates to in-paint the raw depth. iPhone sequences are processed and registered by Polycam pose. Long/short captures of each consumer device are registered with global registration and further refined by COLMAP~\cite{Schonberger_Frahm_2016} bundle adjustment. }
	\label{fig:data_pro}
    \vspace{-0.3cm}
\end{figure*}

\begin{table*}[]

\resizebox{\textwidth}{!}{%
\begin{tabular}{cccccccc}
\toprule
Dataset                                                                   & $N_{room}$ & Device                                                                                                      & RGB-D        & $N_{seq}$ & Resolution                                                                      & \begin{tabular}[c]{@{}c@{}}Pose estimation\\ method\end{tabular}                                   & \begin{tabular}[c]{@{}c@{}}Geometry ground \\truth format\end{tabular} \\ \hline \hline
Tanks\&Temples~\cite{knapitsch2017tanks}                                                           & 4          & \begin{tabular}[c]{@{}c@{}}FARO scanner X330~\cite{faro};\\ Sony A7SM2 camera\end{tabular}                              &              & 4         & 1920×1080                                                                       & COLMAP~\cite{Schonberger_Frahm_2016}                                                                                               & point cloud                                                      \\ \hline
Redwood Scan~\cite{park2017colored}                                                             & 5          & \begin{tabular}[c]{@{}c@{}}FARO scanner X330;\\ Asus Xtion Live camera\end{tabular}                         & $\checkmark$ & 5         & 640×480                                                                         & color ICP~\cite{park2017colored}                                                                                            & point cloud                                                      \\ \hline
ETH3D~\cite{schops2017multi}                                                                     & 7          & \begin{tabular}[c]{@{}c@{}}FARO scanner X330;\\ Nikon D3X DSLR camera;\\ Global-shutter camera\end{tabular} &              & 9         & \begin{tabular}[c]{@{}c@{}}high-res: 6048×4032;\\ low-res: 752×480\end{tabular} & COLMAP~\cite{Schonberger_Frahm_2016}                                                                                              & point cloud                                                      \\ \hline
\begin{tabular}[c]{@{}c@{}}Neural RGB-D \\ synthesis dataset~\cite{Azinovic_Martin-Brualla_Goldman_Niebner_Thies_2022}\end{tabular} & 10         & Synthetic camera                                                                                            & $\checkmark$ & 10        & 640×480                                                                         & BundleFusion~\cite{Dai_Nießner_Zollhöfer_Izadi_Theobalt_2016}                                                                                         & \begin{tabular}[c]{@{}c@{}}Blender~\cite{blender}\\ mesh\end{tabular}           \\ \hline
Mip-NeRF 360~\cite{barron2022mip}                                                             & 1          & Fujifilm X100V camera                                                                                       &              & 1         & 3114×2075                                                                     & COLMAP~\cite{Schonberger_Frahm_2016}                                                                                               & -                                                                \\ \hline
Nerfstudio~\cite{tancik2023nerfstudio}                                                                & 1          & mobile phone                                                                                                &              & 1         &                  994×738                                                               & Polycam~\cite{polycam}                                                                                     & -                                                                \\ \hline
MuSHRoom                                                                 & 10         & \begin{tabular}[c]{@{}c@{}}Faro scanner X130;\\ Azure Kinect v2;\\ iPhone 12 Pro Max\end{tabular}                      & $\checkmark$ & 40        & \begin{tabular}[c]{@{}c@{}}Kinect: 1280×720\\ iPhone: 994×738\end{tabular}    & \multicolumn{1}{l}{\begin{tabular}[c]{@{}l@{}}Kinect: Spectacular AI~\cite{spectacularai} \& COLMAP~\cite{Schonberger_Frahm_2016} \\ iPhone: Polycam~\cite{polycam} \& COLMAP~\cite{Schonberger_Frahm_2016} \end{tabular}} & mesh                                                             \\ \bottomrule
\end{tabular}%
}
\caption{Comparison between 3D reconstruction datasets. We only counted the number of indoor rooms from each datasets. MuSHRoom dataset provides the most indoor scenes captured by multiple sensors.}
\label{tab:cmp_datasets}
\end{table*}

This section first presents the procedures for recording real-world indoor room data using the Kinect, iPhone, and Faro scanner. Then, we describe the post-processing steps we applied to the captured data before the evaluation. Lastly, we highlight the key challenges of the obtained dataset.

\subsection{Data Collection}

To create a diverse dataset, we selected rooms with varying shapes, colors, and indoor objects. 
We have chosen 10 real-world rooms while further details on the selected rooms can be found in the supplementary material.
Prior to recording, we take measures to ensure that any personal privacy concerns are addressed and that the rooms do not reveal any confidential information. 
%
%
During the recording process, we ensured that objects within the rooms remained stationary to maintain consistency across devices and that the objects recorded by each device were in the same position.

\subsubsection{Raw data capturing}

In Figure~\ref{fig:data_pro}, we briefly illustrate the data-capturing pipelines for the three devices.
Comparisons of our dataset with others can be found in Table~\ref{tab:cmp_datasets}. 
Compared with other datasets, MuSHRoom provides the most indoor scenes captured with multiple RGB-D devices and has expert-cleaned reference meshes.
All the raw color/depth images, in-painted depth, the estimated pose and point cloud extracted by Spectacular AI SDK~\cite{spectacularai} and Polycam~\cite{polycam}, and the expert-cleaned reference mesh will be provided for further research.
We use three devices to record each room. A faro scanner is used for high-precision point cloud collection for geometry comparison, and consumer device Azure Kinect and iPhone are used to collect RGB-D sequences.

\noindent\textbf{Kinect.}
We use an Azure Kinect depth camera to get synchronized depth and color images at 30 Hz with a laptop. 
The depth images are captured with a resolution of 512x512, and color images at 1280x720 pixels.
We use the wide FoV mode of the depth camera with 2x2 binning to increase the field of view for better room reconstruction.
Inertial Measurement Unit (IMU) data was recorded at 1.6kHz.
For color image capturing, we fixed the white balance for each room and the auto-exposure for 8 rooms except the sauna and olohuone room, which have large illumination variations inside the room.
During capturing, to increase the possibility of capturing all the details of the room for the long capture, we use a visualization system developed by Spectacular AI SDK~\cite{spectacularai} to inspect the integrity of the reconstructed point cloud extracted from the captured RGB-D images in real-time.
%

When evaluating the novel view synthesis, most of the previous methods~\cite{barron2022mip,tancik2023nerfstudio} select keyframes from the sequences uniformly.
%
%
However, this may not reflect the real case in AR/VR.
It is common in real-world applications for users to first scan an entire room with a device and then wear AR glasses to interact with the environment from random positions and directions.
%
Our goal is to simulate this scenario in order to create a more realistic evaluation method.
We recorded two sequences inside each room. 
For the long capture, we try to include all the parts of the whole scene, and when capturing the short one, we attempt to follow a different motion trajectory.
%

\noindent\textbf{iPhone.}
We use an iPhone 12 pro max to record iPhone data with the Polycam app~\cite{polycam}.
During capture, a UI system provided by Polycam is also used to guarantee all the objects, ceiling, floor, and walls have been covered within one capture video.
Auto-exposure and auto-white balance are used by default.
To ensure the stability of the iPhone, we fix it on a DJI OSMO Mobile 3 handle~\cite{dji}.
Following the same pattern with the Kinect device, we collect the second sequence with the iPhone as a test dataset.


\noindent\textbf{Faro scanner.}
To obtain the geometry ground truth reference mesh of each room, Faro Focus 3D X130 Laser Scanner~\cite{faro} fixed on a tripod is used to collect a high-resolution XYZRGB point cloud.
The reach of the laser ranges from 0.6m to 130m. 
%
We have selected the indoor capture mode, which has a range of more than 10 meters and a ranging noise of 0.15 millimeters.
Each scan was set with 360$^\circ$ horizontal, 170$^\circ$ vertical (-60$^\circ$ to 90$^\circ$) with 1/5 resolution, which takes around 9 minutes to record.
The resolution of each scan is 8192x3414 pixels, with a maximum of 28 million points.
%
We opt for the horizontal weighted metering mode for the camera, which utilizes the light from the horizontal direction to determine the optimal exposure setting. This mode is particularly well-suited for indoor rooms with bright ceiling lights.
%
In order to capture a comprehensive view of the room's interior surface, we perform scanning from 4-5 positions for regular rooms and 7-10 positions for larger rooms.
Each position was strategically selected to maximize the coverage of areas that have not been scanned.


\subsubsection{Post-processing}

\noindent\textbf{Kinect.}
After acquiring the raw data, we used Spectacular AI SDK to extract the 6-degree-of-freedom (6DoF) pose in the OpenCV coordinate system~\cite{opencv}. 
Spectacular AI SDK fuses data from RGB-D cameras and IMU sensors and outputs a robust and accurate 6DoF pose for the keyframes extracted from the whole sequence.
It also exports a reconstructed point cloud registered with multi-view information.
%
To address the issue of raw depth images containing multiple holes with invalid depth values, we utilize the z-buffer~\cite{huang2023boosting} to render depth images from the point cloud and then perform hole in-painting.
%
We perform global registration by using COLMAP~\cite{Schonberger_Frahm_2016} to re-calculate the poses for all the images from the long and short capture with bundle adjustment, and then we re-scale and rotate the COLMAP pose to align with the original Spectacular AI pose.
Pose and point cloud optimized by bundle adjustment are used as \textit{estimated pose} and \textit{estimated point cloud} for reconstruction.


\noindent\textbf{iPhone.}
The raw images have a resolution of 1024×768 and a raw depth of 256×192 pixels.
We use Polycom to extract poses for each keyframe with global optimization.
Then, we use scripts in Nerfstudio~\cite{tancik2023nerfstudio} to pre-process the RGB-D images to get cropped color images as well as up-scaled aligned depth images with a resolution of 994×738 pixels.
To register long and short sequences, we also use COLMAP to re-calculate the COLMAP pose as \textit{estimated pose} and align them to the Polycam pose coordination.


\noindent\textbf{Faro Scanner.}
We register scans captured inside one room with FARO SCENE Software.
To further reduce the size of the point cloud without excessive loss of accuracy, we down-sample 3x for each registered point cloud.
When getting mesh from these high-resolution point clouds, most of the previous datasets~\cite{schops2017multi,park2017colored} utilize Poisson reconstruction~\cite{kazhdan2013screened}, which is not suitable for our scenes with complex objects and high occlusion.
To ensure the quality of the mesh, we use Reality Capture~\cite{capturereality} to triangulate mesh from point clouds.
However, there are still artifacts in the reference mesh due to occlusion and complex reflective surface, which are detrimental to the evaluation.
These artifacts are again manually refined by removing wrong vertices and completing holes in MeshLab~\cite{meshlab,LocalChapterEvents} and Blender~\cite{blender}, and will not contribute to the final comparison.

When comparing different benchmarks, the reconstructed mesh needs to be aligned with the ground truth reference meshes.
We use the estimated point cloud of each room and each device to register the reference mesh automatically. For other pipelines that can expose camera poses, we align the predicted result to the estimated point cloud.
Similar to the alignment procedure proposed in Tanks and Temples~\cite{knapitsch2017tanks}, we first initialize a global scaling and alignment with RANSAC and then refine the registration with color ICP to get the final alignment $\mathbf{T}$. 
%
%

%

\subsection{Challenges of the MuSHRoom dataset}

\begin{figure}[t]
	\centering	 
	\includegraphics[width=\linewidth]{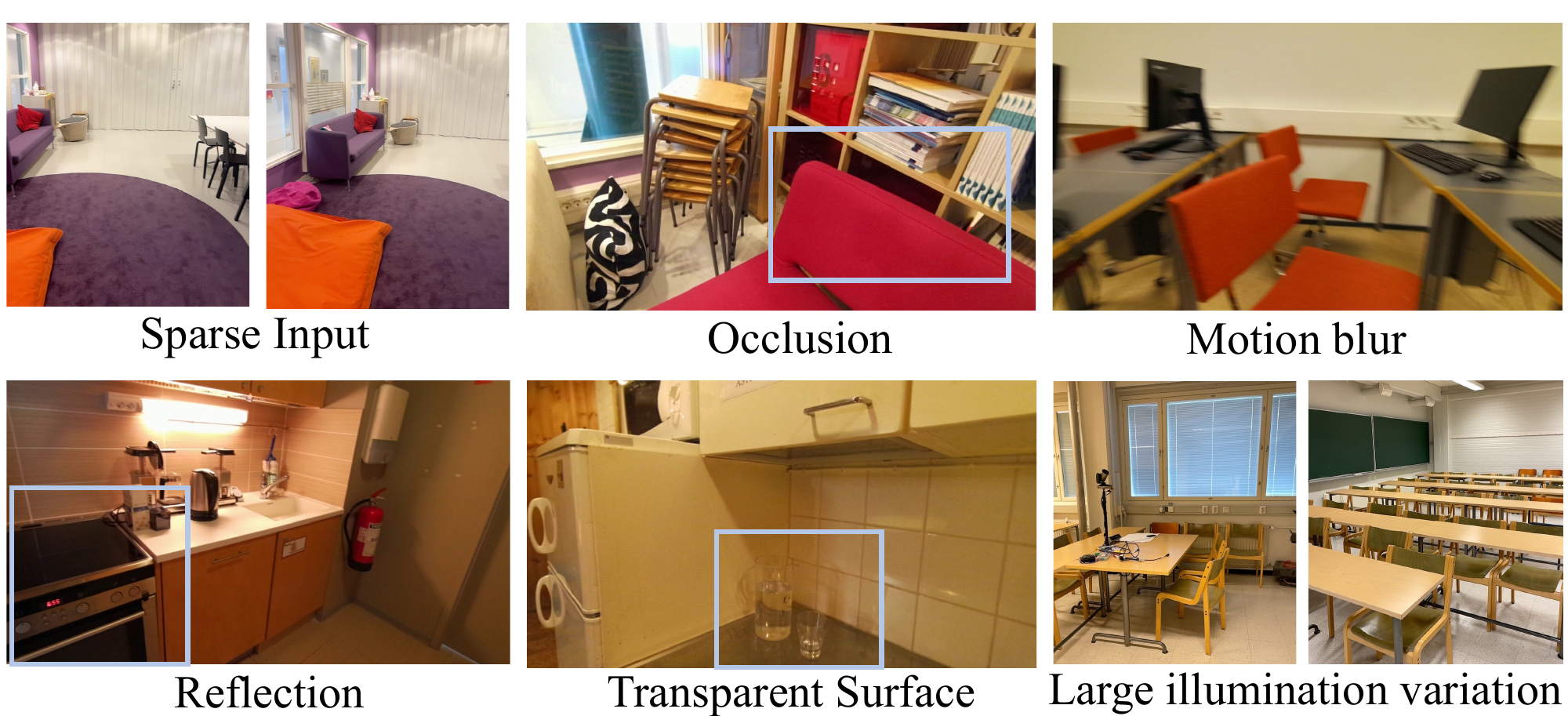}	 	
	\caption{The challenges observed in the MuSHRoom dataset.}
	\label{fig:challenges}
  \vspace{-0.3cm}
\end{figure}

The MuSHRoom dataset introduces several practical challenges, including sparse occlusion, motion blur, reflection, transparent objects, and significant illumination variations, which are detrimental to the training of the reconstruction and rendering models.
In Figure~\ref{fig:challenges}, we illustrate examples of the challenges observed in our dataset.

\noindent\textbf{Sparseness}
To ensure the accuracy of the entire room reconstruction, we only optimize the poses for keyframes with specific view gaps.
%
As a result, the extracted keyframes of each device and room are relatively sparse. This characteristic is not ideal for methods such as Neural Radiance Fields (NeRF), which benefit from dense images as input.

\noindent\textbf{Occlusion}
The layout of objects within each room often includes narrow spaces, making it challenging to capture the backside of many objects. As a result, artifacts can occur during the reconstruction process, as the NeRF models are required to guess the appearance of unseen areas randomly.

\noindent\textbf{Motion Blur}
Unsteady walking patterns and shaky hands can cause images to appear blurry, which will influence the training process.
%

\noindent\textbf{Reflection}
Reflection usually occurs on metal surfaces, like the stove, TV, or mirror, where depth is hard to capture.
The invalid depth is detrimental to the learning for both reconstruction and synthesis tasks.

\noindent\textbf{Transparency}
Transparency is a difficult attribute to learn because of the wrong depth value. These regions are usually completely missing from the mesh model.
%

%

\noindent\textbf{Large illumination variations}
Due to uneven light conditions inside one room, the illumination may vary significantly, making it hard to learn the illumination circumstances and synthesize images as close as possible to the real images.

\noindent\textbf{Evaluation gap}
When training and testing models with different captures and trajectories, the directions and positions of the camera in the training and test set may have large pose differences, which stimulates the pipelines to be robust.
\section{Benchmark}
\label{sec:benchmark}
%

\begin{table*}[]
\caption{Quantitative comparison of different methods in MuSHRoom dataset. We report results average over 10 scenes.}
\resizebox{\textwidth}{!}{%
\begin{tabular}{c|c|ccccc|cccccc}
\toprule
\multirow{3}{*}{Device} & \multirow{3}{*}{Methods} & \multicolumn{5}{c|}{\multirow{2}{*}{Reconstruction quality}}                                                                                   & \multicolumn{6}{c}{Rendering quality}                                                                                                                               \\
                        &                          & \multicolumn{5}{c|}{}                                                                                                                          & \multicolumn{3}{c}{Test within a single sequence}                                & \multicolumn{3}{c}{Test with a different sequence}                               \\
                        &                          & \textbf{Acc} $\downarrow$ & \textbf{Comp}$\downarrow$ & \textbf{C-}$\ell_1$ $\downarrow$ & \textbf{NC} $\uparrow$ & \textbf{F-score}$\uparrow$ & \textbf{PSNR} $\uparrow$ & \textbf{SSIM}$\uparrow$ & \textbf{LPIPS} $\downarrow$ & \textbf{PSNR} $\uparrow$ & \textbf{SSIM}$\uparrow$ & \textbf{LPIPS} $\downarrow$ \\ \midrule
\multirow{4}{*}{iPhone} & Nerfacto~\cite{tancik2023nerfstudio}                 & 0.0652                    & 0.0603                    & 0.0628                           & 0.7491                 & 0.6390                     & 20.83                    & 0.7653                  & 0.2506                      & 20.36                    & 0.7448                  & 0.2781                      \\
                        & Depth-Nerfacto~\cite{tancik2023nerfstudio}           & 0.0653                    & 0.0614                    & 0.0634                           & 0.7354                 & 0.6126                     & 21.23                    & 0.7623                  & 0.2612                      & 20.67                    & 0.7423                  & 0.2873                      \\
                        & MonoSDF~\cite{Yu2022MonoSDF}                & 0.0792                    & 0.0237                    & 0.0514                           & 0.8200                 & 0.7596                     & 19.79                    & 0.6972                  & 0.4122                      & 17.92                    & 0.6683                  & 0.4384                      \\
                        & Splatfacto~\cite{tancik2023nerfstudio}               & 0.1074                    & 0.0708                    & 0.0881                           & 0.7602                 & 0.4405                     & 24.22                    & 0.8375                  & 0.1421                      & 21.39                    & 0.7738                  & 0.1986                      \\ \midrule
\multirow{4}{*}{Kinect} & Nerfacto~\cite{tancik2023nerfstudio}                 & 0.0669                    & 0.0695                    & 0.0682                           & 0.7458                 & 0.6252                     & 23.89                    & 0.8375                  & 0.2048                      & 22.43                    & 0.8331                  & 0.2010                      \\
                        & Depth-Nerfacto~\cite{tancik2023nerfstudio}           & 0.0710                    & 0.0691                    & 0.0701                           & 0.7274                 & 0.5905                     & 24.21                    & 0.8370                  & 0.2107                      & 22.77                    & 0.8345                  & 0.2036                      \\
                        & MonoSDF~\cite{Yu2022MonoSDF}                & 0.0439                    & 0.0204                    & 0.0321                           & 0.8616                 & 0.8753                     & 23.05                    & 0.8315                  & 0.2434                      & 21.60                    & 0.8267                  & 0.2219                      \\
                        & Splatfacto~\cite{tancik2023nerfstudio}               & 0.1007                    & 0.0704                    & 0.0855                           & 0.7689                 & 0.4697                     & 26.07                    & 0.8844                  & 0.1378                      & 23.28                    & 0.8604                  & 0.1579                      \\ \bottomrule
\end{tabular}
}
\label{tab:cmp_new}

\end{table*}
\begin{figure*}[h]
	\centering	 
	\includegraphics[width=\textwidth]{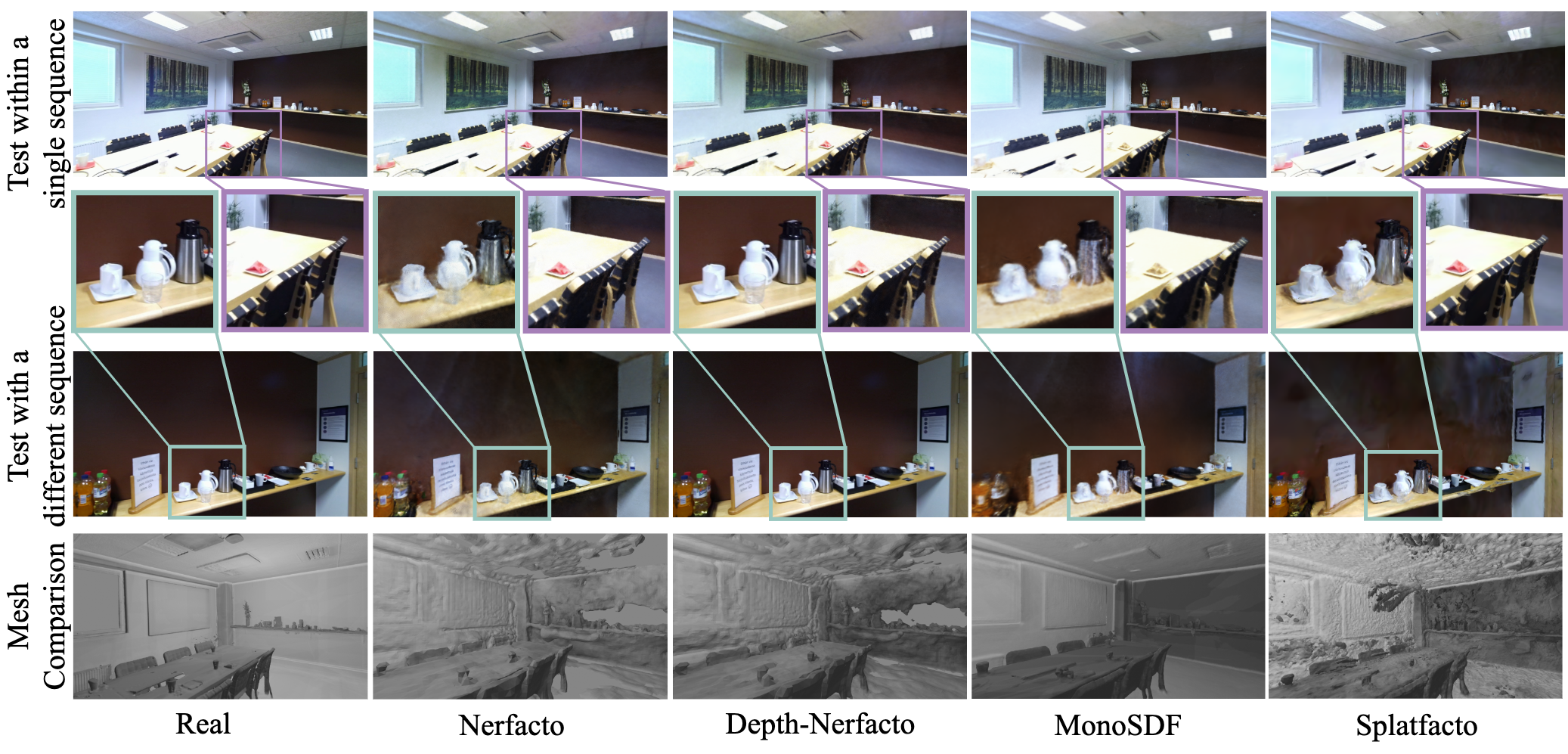}	 	
	\caption{The qualitative comparison of different methods on MuSHRoom dataset. We visualize mesh and novel view synthesis quality.}
\label{fig:kinect_cmp}
\end{figure*}


%


%
%
We propose a new benchmark for developing unified frameworks that focus on realizing both immersive and structurally accurate modeling under real-world constraints. These frameworks are optimized for consumer-grade hardware and operate in an end-to-end fashion.
%
They take as input RGB-D sequences captured by consumer devices and output both accurate 3D mesh models and photorealistic images synthesized from novel views.
We compare the methods for both the mesh reconstruction and novel view synthesis quality quantitatively and qualitatively.
%
%

\subsection{Evaluating Novel View Synthesis}
We test novel view synthesis with two comparison methods: testing within a single sequence and testing with a different sequence.
When comparing with the testing within a single sequence method, we extract keyframes from one sequence as test data and training methods on the other frames of the same sequence.
However, the uniform sampling method usually used in previous methods~\cite{tancik2023nerfstudio,barron2022mip} is not practical in VR/AR applications that require random trajectories.
Thus, we propose a new evaluation protocol, testing with a different sequence, which uses one sequence for training and the other individual sequence for testing. 
The distances and directions of the camera from the same object will be significantly different in the two sequences, which poses a great challenge to the rendering robustness of the pipelines.

\noindent\textbf{Metrics.}
We compare images synthesized from novel views with PSNR, SSIM~\cite{1284395}, and LPIPS~\cite{zhang2018unreasonable} evaluation metrics.

\noindent\textbf{Training and evaluation strategy.}
When testing within a single sequence, we select 10\% frames from the long sequence uniformly as the test set, and others are used as training datasets.
When testing with a different sequence, we use the same model trained with the long capture frames to test the frames in the short capture.

\subsection{Evaluating Geometric Reconstruction}
\label{sec:evaluate}

\noindent\textbf{Metrics.}
We evaluate the predicted mesh models with the reference mesh from both accuracy and completeness aspects.
Following the mesh evaluation protocol introduced in GO-Surf~\cite{wang2022go}, we measure accuracy (Acc), completion (Comp), Chamfer distance (C- $\ell_1$), normal consistency (NC) and F-score metrics when evaluating reconstruction results.
The comparison is conducted between the point cloud sampled from the predicted mesh and reference mesh at a density of 1 point per $cm^2$.
The threshold for computing the F-score is 5 cm.
More details can be seen in the Supplementary Materials.

\noindent\textbf{Mesh Culling.}
In previous methods~\cite{wang2022go,Azinovic_Martin-Brualla_Goldman_Niebner_Thies_2022}, the mesh will be culled according to whether the surfaces have been observed, occluded, or have valid depth before evaluation.
Here we cut the predicted mesh outside the silhouette of the reference mesh.
Because in MuSHRoom some rooms are unbounded or non-square scenes, which cannot be simply culled by square-style bounding boxes and previous assumptions.
%
%
%
We project the predicted mesh and reference mesh to the xy, yz, and xz planes separately and remove vertices and their corresponding triangles of the predicted mesh that out of the contours of the projection of the reference mesh.
After getting a predicted mesh that is not influenced by the outside surface, we only compare the parts owned by reference mesh.
For iPhone sequences, we only apply this cutting method, and the details about the culling strategy can be seen in the Supplementary Materials.

\noindent\textbf{Training and evaluation strategy.}
To compare mesh quality, we utilize training frames from the long capture of each device and room as inputs and compare the resulting mesh with the cleaned ground truth mesh.


\section{Experiments}

%
We compare our baseline with several representative pipelines for both reconstruction and rendering quality, and the detailed instructions for these methods can be found in the Supplementary Materials.

\subsection{Quantitative Evaluation}

We calculate the average metrics of reconstruction and rendering quality for all rooms and show them in Table~\ref{tab:cmp_new}\footnote{We update the results from the previous version of the paper. The previous table can be seen in the supplementary materials}.
%
%
%
Nerfacto~\cite{tancik2023nerfstudio}, Depth-Nerfacto~\cite{tancik2023nerfstudio} and Splatfacto\cite{tancik2023nerfstudio} are excellent in rendering quality but are much worse than SDF-based method in terms of mesh completeness.
MonoSDF~\cite{Yu2022SDFStudio} predict SDF, which regulates the mesh without ambiguity, but its synthesis qualities are worse than NeRF and 3DGS.
%
%
The results also highlight that the inherent complexities of our dataset impede the enhancement of rendering fidelity, which requires further advanced methods developed for real scenarios.

\subsection{Qualitative Evaluation}

We show the qualitative results of testing within one sequence and testing with different sequences, mesh quality of the Kinect sequences in Figure~\ref{fig:kinect_cmp}.
The visualization shows that there is still significant room to get pipelines to perform well for both novel view synthesis and mesh reconstruction.
%


\section{Conclusion}

We have proposed a real-world dataset and a new benchmark with multiple sensors for evaluating pipelines on both 3D reconstruction accuracy and novel view synthesis quality.
The new dataset poses more realistic challenges and supports more practical evaluation.
With consumer-grade devices to collect inputs, pipelines are encouraged to be robust, generalized, and computationally efficient.
We also propose a new method and evaluate it with several popular pipelines, revealing the aim to realize both geometry accuracy and immersion still has a long way to go.
Our dataset can serve as a foundation for the development of a unified framework training in an end-to-end fashion.

\section{Acknowledgement}
The work was supported by the Academy of Finland projects \#324346 and \#353139, and also carried out with the support of Centre for Immersive Visual Technologies (\href{https://civit.fi/}{CIVIT}) research infrastructure, Tampere University, Finland.

{\small
\bibliographystyle{ieee_fullname}
\bibliography{egbib}
}

\end{document}


\newcommand{\myicon}{\includegraphics[scale=0.03]{3DV2022AuthorKit/Figure/mushroom.png}}

\title{Supplementary Material of MuSHRoom: Multi-Sensor Hybrid Room \\ Dataset for Joint 3D Reconstruction and Novel View Synthesis}

\author{Xuqian Ren,$^{1}$  Wenjia Wang,$^{2}$  Dingding Cai,$^{1}$  Tuuli Tuominen,$^{1}$
 Juho Kannala,$^{3}$  Esa Rahtu$^{1}$  \\
{}$^1$Tampere University, Finland {}$^2$The University of Hong Kong, China {}$^3$Aalto University, Finland\\
{\tt\small \{xuqian.ren,dingding.cai,tuuli.tuominen,esa.rahtu\}@tuni.fi wwj2022@connect.hku.hk } \\
{\tt\small Juho.Kannala@aalto.fi }
}

\maketitle



\section{Dataset details}
\label{sec:rationale}
%
\subsection{Visualization System during capture}

\begin{figure}[h]
	\centering	 
	\includegraphics[width=\columnwidth]{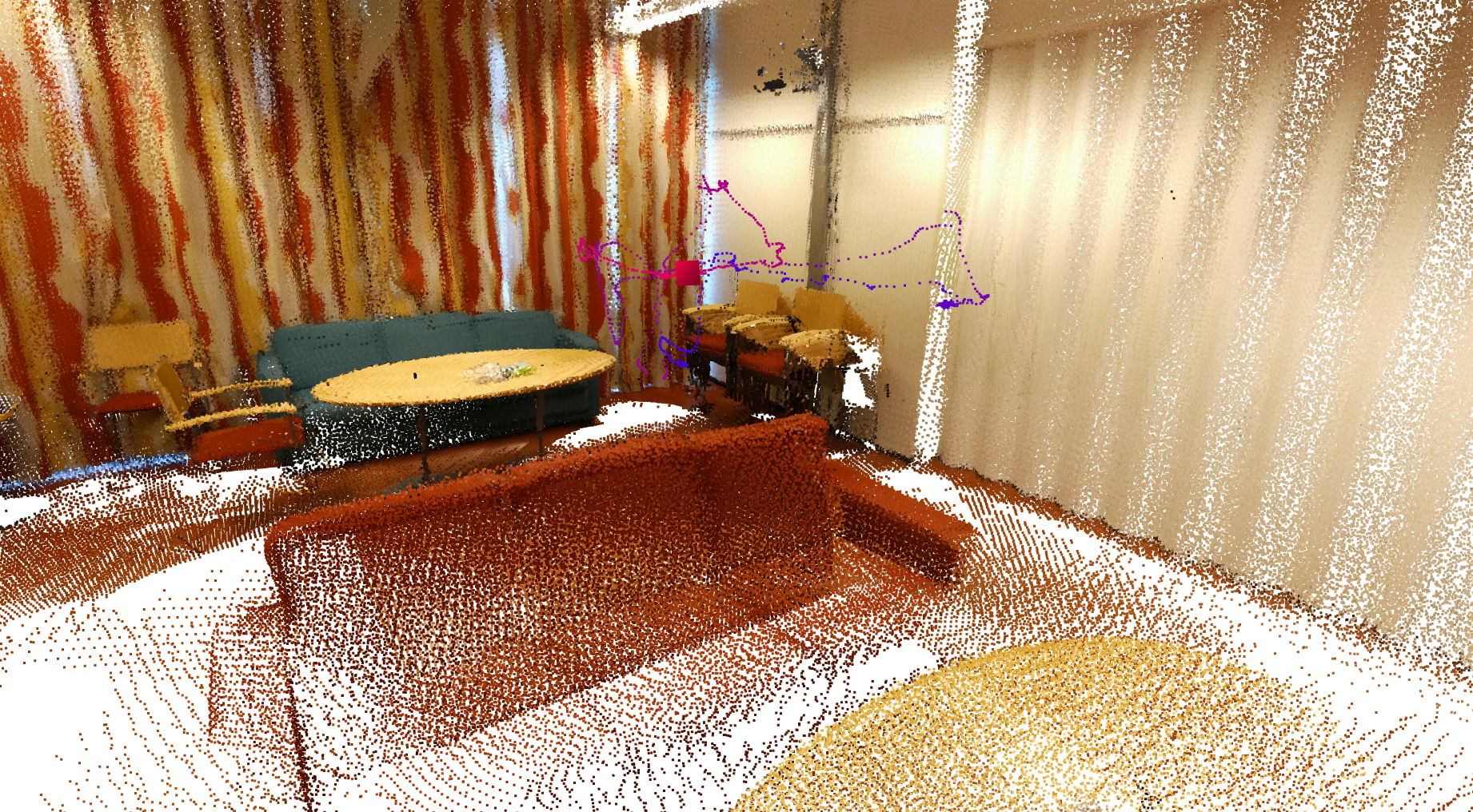}	 	
	\caption{The visualization system. 
 The colored area indicates successful capture, while the empty area is repeatedly scanned 
 by checking the integrity of the point cloud.}
	\label{fig:visualization}
\end{figure}

We leverage a visualization system developed by Spectacular AI SDK~\cite{spectacularai} to inspect the integrity of the point cloud reconstructed from the captured RGB-D images in real-time when using Kinect.
In Figure~\ref{fig:visualization}, we show an example of the visualization interface. 
%

\subsection{The details of each room}

\begin{figure*}[t]
	\centering	 
	\includegraphics[width=\textwidth]{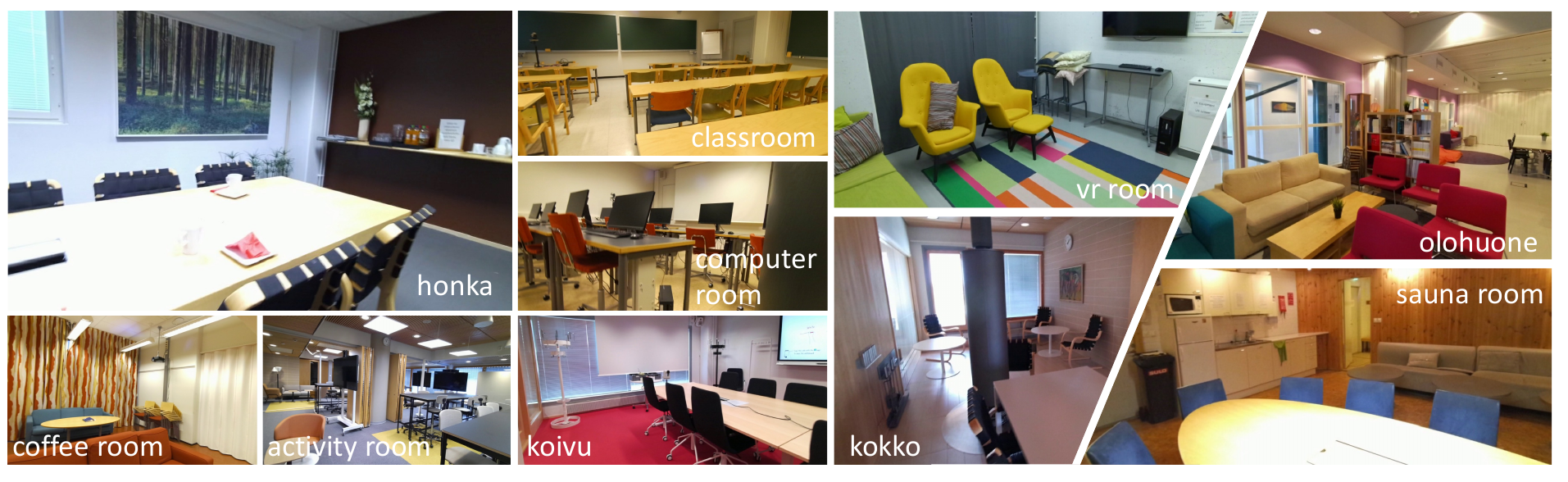}	 	
	\caption{The MuSHRoom dataset. Our dataset contains ten rooms with different shapes, colors, illumination, and objects. We show an image example of each room captured by Kinect.}
	\label{fig:exp}
\end{figure*}

In Figure~\ref{fig:exp}, we show an example image of each room. The MuSHRoom is a room-scale dataset with various styles, colors, illumination, and objects, demonstrating real-world challenges.
In Table~\ref{stats:data}, we show the details of the captured rooms, including the room names, scales, and camera settings.

\begin{table}[]

\resizebox{\columnwidth}{!}{%
\begin{tabular}{ccccccc}
\hline
Scene         &  Scale (m)       & \begin{tabular}[c]{@{}c@{}}Exposure \\ time \\ (µs)\end{tabular} & \begin{tabular}[c]{@{}c@{}}White \\ Balance \\ (K)\end{tabular} & Brightness & Gain \\ \hline
coffee room   &  6.3 $\times$ 5 $\times$ 3.1   & 41700                                                           & 2830                                                         & 128        & 130  \\
computer room &  9.6 $\times$ 6.1 $\times$ 2.5 & 33330                                                           & 3100                                                         & 128        & 255  \\
classroom     &  8.9 $\times$ 7.2 $\times$ 2.8 & 33330                                                           & 3300                                                         & 128        & 88   \\
honka         &  6.1 $\times$ 3.9 $\times$ 2.3 & 16670                                                           & 3200                                                         & 128        & 128  \\
koivu         &  10 $\times$ 8 $\times$ 2.5    & 16670                                                           & 4200                                                         & 128        & 128  \\
vr room       &  5.1 $\times$ 4.4 $\times$ 2.8 & 8300                                                            & 3300                                                         & 128        & 88   \\
kokko         &  6.7 $\times$ 6.0 $\times$ 2.5 & 133330                                                          & 4000                                                         & 128        & 255  \\
sauna         &  9.9 $\times$ 6.5 $\times$ 2.4               & Auto                                                            & 3300                                                         & Auto       & Auto \\
activity      &  12 $\times$ 9 $\times$ 2.5    & 50000                                                           & 3200                                                         & 128        & 130  \\
olohuone      &  19 $\times$ 6.4 $\times$ 3    & Auto                                                            & 3600                                                         & Auto       & Auto \\ \hline
\end{tabular}%
}
\caption{The parameters of each room in our datasets. We introduce the room names, room scales and camera parameters. }
\label{stats:data}
\end{table}
When using COLMAP~\cite{Schonberger_Frahm_2016} to calculate the globally optimized pose for activity and koivu room captured with iPhone sequence, COLMAP failed to calculate accurate poses, so we walked around twice and captured a long sequence.
We cut the long sequence from the middle and use the original Polycom of the first circle for training and the frames in the second circle for testing.

\section{Comparison methods}
This section introduces the baseline methods we have compared with.

\noindent\textbf{Nerfacto:}
Nerfstudio~\cite{tancik2023nerfstudio} is an end-to-end workflow that encapsulates various state-of-the-art NeRF techniques, which is friendly to user-collected real-world data.
Nerfacto is one of the NeRF pipelines assembled by Nerfstudio that combines components from practical novel methods to balance speed and quality.
With a proposal sampler~\cite{barron2022mip}, scene contraction~\cite{barron2022mip}, and density field, Nerfacto can achieve immersive novel view synthesis quality even with real-world noisy data.
However, the density field is optimized exclusively for visual consistency, which means that this model sacrifices geometry accuracy and creates occupancy regions to support the volumetric rendering even in parts of the space that are not occupied by the underlying surface. 
The surface will be predicted with the help of the density rather than accurate zero thickness surface~\cite{rakotosaona2023nerfmeshing}.
When the surface is not sharp enough, the consistency of the predicted depth and normal from multiple views cannot be guaranteed, leading the 3D model extracted from the density field to become ambiguity~\cite{xiao2022resnerf} when representing mesh with truncated signed distance function (TSDF).
In our evaluation, we use Poisson surface reconstruction~\cite{Kazhdan_Bolitho_Hoppe_2006} to extract the mesh model.

\noindent\textbf{Depth-Nerfacto:}
Depth-Nerfacto~\cite{tancik2023nerfstudio} is the method to add depth supervision on Nerfacto with the sensor depth. We follow the default setting in Nerfstudio, and use the depth loss proposed in DS-NeRF~\cite{kangle2021dsnerf}.

\noindent\textbf{MonoSDF:}
SDFStudio~\cite{Yu2022SDFStudio} is a unified framework that focuses on 3D reconstruction based on Nerfstudio, combined with recent techniques designed from implicit surface reconstruction.
Similar to Nerfacto, we choose the re-implementation of MonoSDF~\cite{Yu2022MonoSDF} in SDFStudio and keep the default setting.
SDF output can largely improve the geometry accuracy, but it constrains the occupancy predicting flexibility, which impedes the learning of details in appearance during volume rendering~\cite{xiao2022resnerf}.
Its mesh and novel view synthesis also be easily influenced by the pose pre-processing. If the poses are not re-scaled inside the unit properly, the extracted mesh and synthesized images will be impacted.

\noindent\textbf{Splatfacto:}
We use Splatfacto~\cite{tancik2023nerfstudio}, a 3D Gaussian splatting (3DGS) method re-implemented by Nerfstudio, to test 3DGS on our dataset. We keep the default setting in Nerfstudio. We use Poisson surface reconstruction to extract the mesh model.
%


\section{Experiments}

\subsubsection{Implementation Details:}
We train each method 30k iterations. 
When training Nerfacto, Depth-Nerfacto and NeuS-facto, we did not include camera pose optimization.
Other settings are the same as the default setting reported in each paper.
%

\subsection{Metrics}

\noindent\textbf{Metrics for comparing reconstruction}
We compare the mesh reconstruction ability from both the accuracy and completeness aspects.
As introduced in~\cite{wang2022go}, we measure accuracy (Acc), completion (Comp), Chamfer distance (C- $\ell_1$), normal consistency (NC), and F-score metrics when evaluating reconstruction results. 
Acc refers to what proportion of the predicted point cloud aligns with a reference point cloud with a certain threshold.
Comp refers to how well a reconstructed mesh represents the full reference mesh. 
C- $\ell_1$ distance measures the similarity between the predicted point cloud and the ground truth point cloud. It computes the average distance from a point in one point cloud to the nearest point in the other point cloud, measuring how closely two point clouds are in space.
NC refers to the alignment of normals between two surfaces, representing the influence of the orientation of the surface.
F-score used to balance the percison $P$ and recall $R$ by
\begin{equation}
F_{score} = \frac{2PR}{P+R}
\end{equation}

Precision comes from the percentage of Acc within a threshold:
\begin{equation}
P\left(t_i\right)=\frac{1}{n} \sum_{j=1}^n I\left(Acc_j \leq t_i\right)
\end{equation}
Recall comes from the percentage of Comp within a threshold:
\begin{equation}
R\left(t_i\right)=\frac{1}{n} \sum_{j=1}^n I\left(Comp_j \leq t_i\right)
\end{equation}
In our comparison, we set the threshold $t_i$ to 5cm.

\noindent\textbf{Metrics for comparing novel view synthesis}

We use PSNR, SSIM~\cite{1284395}, and LPIPS~\cite{zhang2018unreasonable} to mesh the pixel and feature distances between synthesized images and real images.

\begin{figure}[t]
	\centering	 
	\includegraphics[width=\columnwidth]{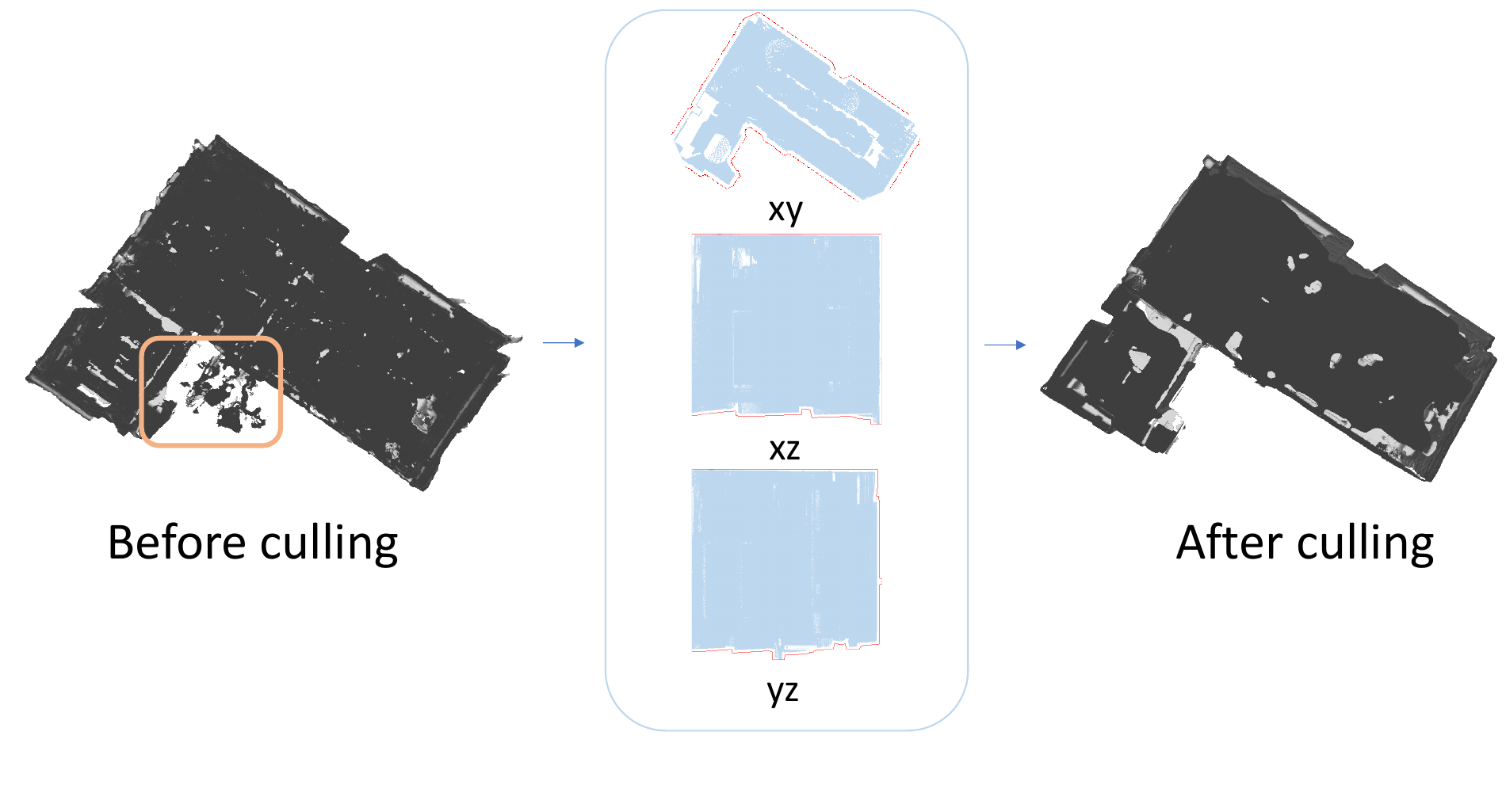}	 	
	\caption{Visualization of our culling protocol. On the left is a predicted mesh of the "koivu" room, reconstructed from a Kinect sequence and culled using the prior culling protocol. Notably, there remains redundant mesh, as highlighted in the yellow square. However, when culled based on the contour of the reference mesh's projections, the mesh is cleanly trimmed.}
	\label{fig:cull}
\end{figure}


\begin{table*}[htb]
\resizebox{\textwidth}{!}{%
\begin{tabular}{ccccccccccccc}
\hline
\multirow{3}{*}{Device} & \multirow{3}{*}{Methods} & \multicolumn{5}{c}{\multirow{2}{*}{Reconstruction quality}}                             & \multicolumn{6}{c}{Rendering quality}                                                                   \\ \cline{8-13} 
                        &                          & \multicolumn{5}{c}{}                                                                    & \multicolumn{3}{c}{Test within a single sequence}  & \multicolumn{3}{c}{Test with a different sequence} \\ \cline{3-13} 
                        &                          & \textbf{Acc} $\downarrow$             & \textbf{Comp}$\downarrow$            & \textbf{C-}$\ell_1$ $\downarrow$            & \textbf{NC} $\uparrow$              & \textbf{F-score} $\uparrow$         & \textbf{PSNR} $\uparrow$           & \textbf{SSIM}  $\uparrow$            & \textbf{LPIPS} $\downarrow$           & \textbf{PSNR} $\uparrow$           & \textbf{SSIM}  $\uparrow$            & \textbf{LPIPS} $\downarrow$           \\ \hline
\multirow{5}{*}{Kinect} & Volumetric Fusion~\cite{curless1996volumetric}         & 0.0354          & 0.0341          & 0.0347          & 0.8159          & 0.8439          & 13.84          & 0.6628          & 0.4208          & 13.10          & 0.6509          & 0.4423          \\
                        & GO-Surf~\cite{wang2022go}                  & 0.0355          & 0.0367          & 0.0361          & 0.8664          & 0.8620          & 20.56          & 0.7708          & 0.3095          & 20.01          & 0.7693          & 0.2812          \\
                        & Nerfacto~\cite{tancik2023nerfstudio}                 & 0.0570          & 0.1485          & 0.1027          & 0.6878          & 0.5715          & \cellcolor[HTML]{FFD966}{22.08}          & \cellcolor[HTML]{FFD966}{0.7971}          & \cellcolor[HTML]{FFC7CE}{0.2479} & \cellcolor[HTML]{FFC7CE}{22.60} & \cellcolor[HTML]{FFC7CE}{0.8457} & \cellcolor[HTML]{FFC7CE}{0.1822} \\
                        & NeuS-facto~\cite{Yu2022SDFStudio}               & \cellcolor[HTML]{FFC7CE}{0.0294} & \cellcolor[HTML]{FFC7CE}{0.0253} & \cellcolor[HTML]{FFC7CE}{0.0274} & \cellcolor[HTML]{FFC7CE}{0.8738} & \cellcolor[HTML]{FFC7CE}{0.9025} & 20.61          & 0.7799          & 0.2760          & 21.50          & 0.8285          & 0.2094          \\
                        & Ours                     & \cellcolor[HTML]{FFD966}{0.0295}          & \cellcolor[HTML]{FFD966}{0.0258}          & \cellcolor[HTML]{FFD966}{0.0276}          & \cellcolor[HTML]{FFD966}{0.8736}          & \cellcolor[HTML]{FFD966}{0.8995}          & \cellcolor[HTML]{FFC7CE}{22.26} & \cellcolor[HTML]{FFC7CE}{0.8040} & \cellcolor[HTML]{FFD966}{0.2608}          & \cellcolor[HTML]{FFD966}{22.45}          & \cellcolor[HTML]{FFD966}{0.8423}          & \cellcolor[HTML]{FFD966}{0.2012}          \\ \hline
\multirow{5}{*}{iPhone} & Volumetric Fusion~\cite{curless1996volumetric}         & \cellcolor[HTML]{FFC7CE}{0.0521} & \cellcolor[HTML]{FFC7CE}{0.0207} & \cellcolor[HTML]{FFC7CE}{0.0364} & 0.7867          & \cellcolor[HTML]{FFC7CE}{0.8050} & 11.74          & 0.5519          & 0.5126          & 11.80          & 0.5525          & 0.5094          \\
                        & GO-Surf~\cite{wang2022go}                  & 0.0630          & \cellcolor[HTML]{FFD966}{0.0305}          & \cellcolor[HTML]{FFD966}{0.0468}          & \cellcolor[HTML]{FFC7CE}{0.8401} & \cellcolor[HTML]{FFD966}{0.7818}          & 17.50          & 0.6292          & 0.4620          & 17.64          & 0.6247          & 0.4832          \\
                        & Nerfacto~\cite{tancik2023nerfstudio}                 & \cellcolor[HTML]{FFD966}{0.0592}          & 0.1450          & 0.1021          & 0.6661          & 0.5973          & \cellcolor[HTML]{FFC7CE}{20.53} & \cellcolor[HTML]{FFC7CE}{0.7555} & \cellcolor[HTML]{FFC7CE}{0.2696} & \cellcolor[HTML]{FFC7CE}{20.72} & \cellcolor[HTML]{FFC7CE}{0.7625} & \cellcolor[HTML]{FFC7CE}{0.2670} \\
                        & NeuS-facto~\cite{Yu2022SDFStudio}               & 0.0659          & 0.0453          & 0.0556          & 0.8135          & 0.7200          & 17.11          & 0.6696          & 0.4501          & 16.83          & 0.6504          & 0.4608          \\
                        & Ours                     & 0.0629          & 0.0448          & 0.0539          & \cellcolor[HTML]{FFD966}{0.8231}          & 0.7281          & \cellcolor[HTML]{FFD966}{19.29}          & \cellcolor[HTML]{FFD966}{0.7130}          & \cellcolor[HTML]{FFD966}{0.3898}          & \cellcolor[HTML]{FFD966}{18.29}          & \cellcolor[HTML]{FFD966}{0.6819}          & \cellcolor[HTML]{FFD966}{0.3985}          \\ \hline
\end{tabular}%
}
\caption{The average metrics of reconstruction and rendering quality for all rooms. The best results are highlighted in pink. The second best results are marked in yellow. Test within a single sequence means we uniformly sample test frames from a single sequence and train on all left frames. Test with a different sequence means we train on one sequence and test on another individual sequence.}
\label{cmp:all}
\vspace{-0.4cm}
\end{table*}

\subsection{Mesh culling protocol}
In the previous method~\cite{wang2022go}, the mesh is culled based on several criteria: first subdivided to have the maximum edge length below 1.5cm and then culled by whether the parts are visible within the camera's frustum, and if there's valid depth in the corresponding region, and if they are occluded.
However, we noticed that for some non-rectangular rooms without precise boundaries, not all redundant mesh parts are effectively culled.
For instance, as depicted in Figure~\ref{fig:cull}, the mesh takes on an "L" shape. The exterior mesh is not culled because it can be observed through a transparent window door. 
Therefore, meshes culled using the previous protocol can lead to imprecise comparison results.
Here, we propose a new culling method that uses the boundary of the projection of the reference mesh to further cull the predicted mesh.
In our culling protocol, after aligning the predicted mesh to the reference mesh, we project the reference mesh into the xy, xz, and yz planes.
To avoid the boundaries being too close to the predicted mesh and causing some incorrect cuts, we first dilate the projections.
Then we detect the contours of three projections and cut the parts of the predicted mesh that are outside of the contours.
We only employed our culling protocol in the mesh evaluation. 
We compare all regions culled by our protocol with the reference mesh.

\subsection{Quantitative Results when using none global optimized pose}
The table we reported in the first version of our paper uses different poses when evaluating the "Test within a single sequence" and "Test with a different sequence". The poses used for "Test within a single sequence" in previous results are got from spectacularAI pose/Polycam pose for Kinect/iPhone. The previous training and testing method needed to train two times, we abandoned this method for efficiency. But for clarification, we also report the previous results there ~\cref{cmp:all}.

{\small
\bibliographystyle{ieee_fullname}
\bibliography{egbib}
}